\documentclass[lettersize,journal]{IEEEtran}
\usepackage{amsmath,amsfonts}
\usepackage{algorithmic}
\usepackage{algorithm}
\usepackage{array}
\usepackage[caption=false,font=normalsize,labelfont=sf,textfont=sf]{subfig}
\usepackage{textcomp}
\usepackage{stfloats}
\usepackage{url}
\usepackage{verbatim}
\usepackage{graphicx}
\usepackage{cite}
\usepackage{multirow}
\usepackage{makecell}
\usepackage{bbding}
\usepackage{bm}
\usepackage{xcolor}
\usepackage{color}
\usepackage[table]{xcolor}

\begin{document}

%\title{An Efficient Group and Interactive Transformer with Local Attentive Hashing for Point Cloud Registration}
\title{LAHNet: Local Attentive Hashing Network for Point Cloud Registration}

\author{Wentao Qu, Xiaoshui Huang and Liang Xiao~\IEEEmembership{}
        % <-this % stops a space
%\thanks{This work was supported in part by the Jiangsu Geological Bureau ResearchProject under Grant 2023KY11, in part by the National Natural Science Foundation of China under Grant 61871226 (Corresponding authors: Xiaoshui Huang; Liang Xiao).}        
%\thanks{Wentao Qu are with the Nanjing University of Science and Technology, NanJing 210094, China (e-mail: quwentao@njust.edu.cn)}

\thanks{This work was supported in part by the Jiangsu Geological Bureau Research Project under Grant 2023KY11, in part by the National Natural Science Foundation of China under Grant 62471235., and in part by the Frontier Technologies R\&D Program of Jiangsu under grant BF2024070. (Corresponding authors: Xiaoshui Huang; Liang Xiao).}    
%\thanks{Wentao Qu are with the Nanjing University of Science and Technology, NanJing 210094, China (e-mail: quwentao@njust.edu.cn)}
\thanks{Wentao Qu is with the Nanjing University of Science and Technology, NanJing 210094, China (e-mail: quwentao@njust.edu.cn)}
\thanks{Xiaoshui Huang is with the Shanghai Jiao Tong University, Shanghai 200030, China (e-mail: huangxiaoshui@163.com).}
\thanks{Liang Xiao is with the Nanjing University of Science and Technology, NanJing 210094, China (e-mail: xiaoliang@mail.njust.edu.cn)}
%\thanks{Abiao Li is the Jiangxi University of Finance and Economics, Jiangxi 330013, China (e-mail: lab183@foxmail.com).}
}

\markboth{Journal of \LaTeX\ Class Files,~Vol.~14, No.~8, August~2021}%
{Shell \MakeLowercase{\textit{et al.}}: A Sample Article Using IEEEtran.cls for IEEE Journals}


%significantly reducing the computational cost of partitioning windows

%In this context, LAHNet can accept large-scale point clouds as input, learning robust and distinctive features.

% The small receptive field leads to feature ambiguity on ambiguous geometric structures

% In contrast, the large receptive field significantly improving the understanding of features regarding geometric information, enhancing feature distinctiveness. 

\maketitle

\begin{abstract}
Most existing learning-based point cloud descriptors for point cloud registration focus on perceiving local information of point clouds to generate distinctive features. However, a reasonable and broader receptive field is essential for enhancing feature distinctiveness. In this paper, we propose a \textbf{L}ocal \textbf{A}ttentive \textbf{H}ashing \textbf{Net}work for point cloud registration, called LAHNet, which introduces a local attention mechanism with the inductive bias of locality of convolution-like operators into point cloud descriptors. Specifically, a Group Transformer is designed to capture reasonable long-range context between points. This employs a linear neighborhood search strategy, Locality-Sensitive Hashing, enabling uniformly partitioning point clouds into non-overlapping windows. Meanwhile, an efficient cross-window strategy is adopted to further expand the reasonable feature receptive field. Furthermore, building on this effective windowing strategy, we propose an Interaction Transformer to enhance the feature interactions of the overlap regions within point cloud pairs. This computes an overlap matrix to match overlap regions between point cloud pairs by representing each window as a global signal.  Extensive results demonstrate that LAHNet can learn robust and distinctive features, achieving significant registration results on real-world indoor and outdoor benchmarks.

%Extensive results on real-world indoor and outdoor benchmarks demonstrate that LAHNet enables to generate distinctive point cloud descriptors, achieving significant registration results.
\end{abstract}


% 3D reconstruction \cite{wang2020computational}, 

% \cite{greve2024collaborative} \cite{vinodkumar2024deep}
 
\section{Introduction}
\IEEEPARstart{G}{iven} two partially overlapped point cloud fragments, the goal of point cloud registration is to align them in the same coordinate system by estimating a transformation matrix. It is vital to many downstream tasks, such as autonomous driving , robotics technology and 3D reconstruction  \cite{qu2025robust, qu2025self}. Within existing registration methodologies, descriptor-based methods constitute a pivotal category \cite{zeng20173dmatch, choy2019fully, huang2021predator, ao2021spinnet, huang2021imfnet, poiesi2022learning, wang2023roreg, yu2024riga}, attaining state-of-the-art accuracy on extensive real-world datasets. The performance of these descriptor-based methods is contingent upon the feature distinctiveness \cite{huang2021predator, huang2021imfnet, poiesi2022learning, wang2023roreg, yu2024riga}.

\begin{figure}[htp]
	\centering
	\includegraphics[width=0.48\textwidth]{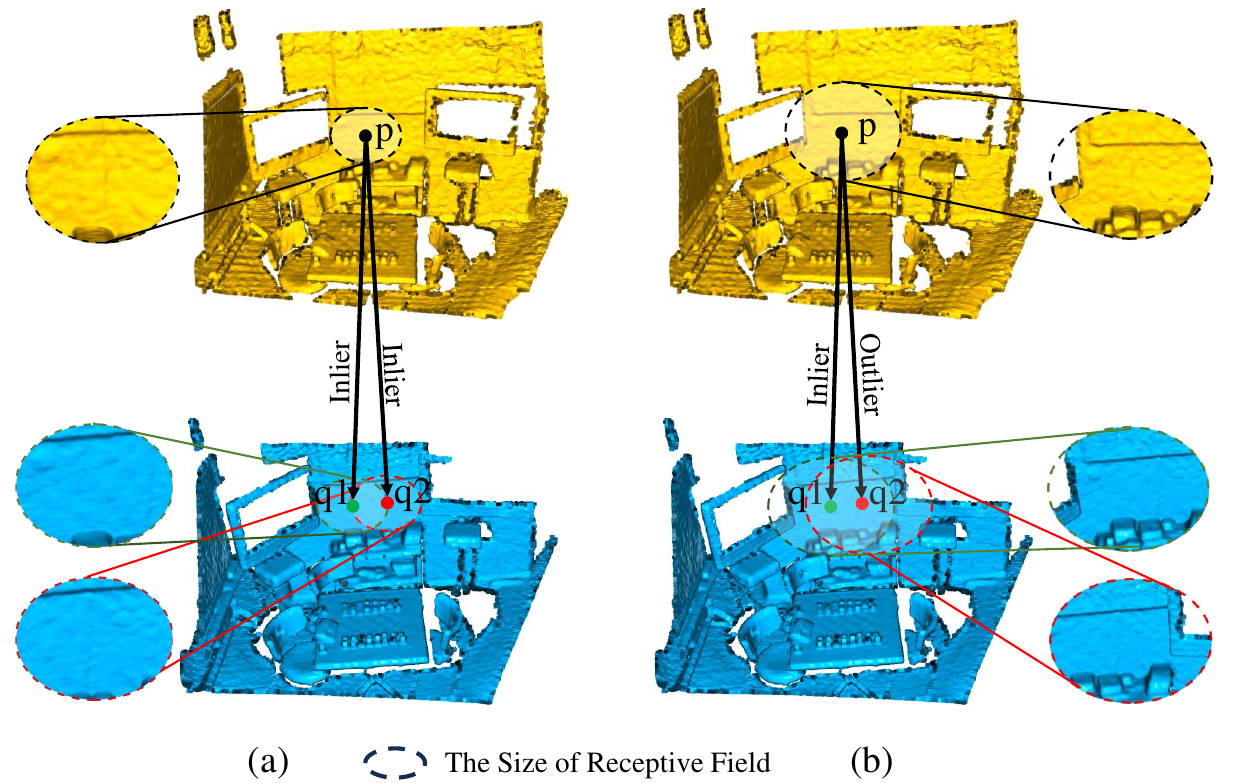}
    \vspace{-10pt}
	\caption{Based on the corresponding position on point cloud pairs, $\bm{p}$ corresponds to $\bm{q^1}$  (an inlier) correctly, while $\bm{p}$ corresponds to $\bm{q^2}$ (an outlier) incorrectly. (a) The limited receptive field results in very similar geometric representation between $\bm{q^1}$ and $\bm{q^2}$, may letting $\bm{p}$ incorrectly match $\bm{q^2}$ as an inlier. (b) The reasonable and expansive (non-global) receptive field enables features to more comprehensively perceive geometric information, making $\bm{q^1}$ and $\bm{q^2}$ express different geometric information. This significant feature distinctiveness enables $\bm{p}$ to correctly match $\bm{q^2}$ as an outlier.}
	\label{fig_example}
\end{figure}

Inspired by deep learning, the representative work FCGF \cite{choy2019fully} employs a fully convolutional network to learn local features, significantly surpassing hand-crafted descriptors for feature distinctiveness. Building upon this pattern, many learning-based descriptor-based methods focus on optimize local feature extraction \cite{zeng20173dmatch, huang2021predator, ao2021spinnet, poiesi2022learning, wang2023roreg}. However, while these methods have achieved improved results, they fail to model reasonable long-range dependencies. Fig \ref{fig_example} illustrates the importance of reasonable long-range dependencies for enhancing feature distinctiveness. The limited receptive field results in feature ambiguity, particularly on geometric structures with similar local patterns such as the wall and the floor\cite{huang2021imfnet}. In contrast, the expansive (non-global) receptive field significantly enhances features distinctiveness by improving the understanding of point surrounding geometric information.

%  the quadratic computation complexity for input points leads to only limited attempts to apply Transformer to 3D point clouds

Along another research avenue, Transformer \cite{vaswani2017attention} can naturally provide long-range signals through the self-attention mechanism \cite{guo2021pct}. However, some studies suggest that \textbf{the global receptive field may introduce redundant disturbing information \cite{xia2022vision} and cause feature averaging \cite{park2022how}}. Moreover, the quadratic computation complexity limits applications of Transformer in point clouds \cite{mao2021voxel, qu2024conditional}. It is worth noting that Swin Transformer \cite{liu2021swin} adopting the local attention mechanism using shifted windows, has shown remarkable results in image tasks. Swin Transformer first linearly partitions regular and ordered images into non-overlapping windows, and then performs self-attention within each window using regular and offset patterns. This introduces the inductive bias of locality of convolution-like operators, and enables a significant reduction in memory and computational complexity.

% , and it is challenging to handle the last window reasonably

Inspired by the above, to introduce the local attention mechanism into point clouds, a straightforward strategy is to partition disordered point clouds into regular windows without overlap using voxelization or KNN \cite{fan2022embracing, qu2025end}. However, this often leads to significant memory and computation waste due to numerous empty voxels or faces the quadratic complexity of neighborhood searches. Recently, some methods have utilized octrees for uniform window partitioning to incorporate local attention mechanisms into 3D tasks, achieving remarkable results \cite{wang2023octformer}.

% Inspired by the above, some methods try to partition the point cloud into regular windows with voxelization \cite{fan2022embracing}. However, this may result in a significant waste of memory and computation, due to the presence of a large number of empty voxels \cite{liu2023flatformer, zhou20213d}. Another approach is to partition the point cloud using KNN \cite{zhao2021point}. But this suffers from quadratic computation complexity for searching neighbors. Recently, some methods utilizing octrees and grid pooling for windows partitioning have achieved remarkable results \cite{wang2023octformer, wu2022point, wu2024point}.

%In this paper, we propose a novel point cloud descriptor, named LAHNet, which utilizes the local attention mechanism to capture long-range dependencies and is memory-efficient. 

% , significantly reducing the computational overhead introduced by partitioning windows.

Unlike previous window partitioning strategies, in this paper, we attempt to uniformly partition point clouds using locality-sensitive hashing (LSH) \cite{andoni2015practical} with linear costs. We propose a \textbf{L}ocal \textbf{A}ttentive \textbf{H}ashing \textbf{Net}work, named LAHNet,  introducing a local attention mechanism with the inductive bias of locality  of convolution-like operators into point cloud descriptors for point cloud registration. Specifically, we design a Group Transformer (GT), which models reasonable long-range dependencies between points to expand the feature receptive field, enhancing feature distinctiveness. To efficiently partition windows, GT uses a simple and effective neighborhood search strategy, LSH, to uniformly partition point clouds into non-overlapping windows. LSH applies a linear function to project point coordinates into corresponding hash values. The close hash values in the hashing domain correspond to the adjacent points in Euclidean space, thereby ensuring the logically reasonable partition for each window.  Meanwhile, GT follows the window shifted mechanism of Swin Transformer, adopting a efficient cross-window interaction strategy to further improve the reasonable feature receptive field.

%Specifically, unlike the previous windowing strategy (voxelization, KNN, octrees and grid pooling), we propose a Group Transformer, which employs locality-sensitive hashing (LSH) \cite{andoni2015practical} to partition 3D space into non-overlapping windows, simply and efficiently introducing the local attention mechanism into point cloud descriptors to establish reasonable long-range dependencies between points. In contrast to utilizing KNN \cite{zhao2021point}, LSH is an approximately linear search strategy in Euclidean space. This first applies a linear function to project point coordinates into hash values. Then, the close hash values correspond to nearby points in Euclidean space, thereby preserving a logically reasonable partition for each window. Simultaneously, we adopt a cross-window interaction strategy. This randomly combines the key and value with points from other windows, further expanding the feature receptive field of the current window by incorporating overlay information. 

Furthermore, to enhance the information exchange of the overlap regions within point cloud pairs,  we propose a Interaction Transformer (IT), which is formally based on multi-head cross-attention. IT first encodes each window as a global signal to compute reasonable matching scores. Then, the overlap matrix is obtained using these matching scores to match overlapping windows. Subsequently, the cross-attention mechanism is applied to the matched window pairs, enhancing the feature distinctiveness on low-overlap point cloud pairs.

In this way, LAHNet can achieve significant registration results on high-overlap (overlap ratio $>30\%$) \cite{zeng20173dmatch}, low-overlap (overlap ratio $10\%$-$30\%$) \cite{huang2021predator} and outdoor \cite{geiger2012we} point cloud pairs. Our key contributions are as follows:

\begin{itemize}
	\item We design the Group Transformer, which leverages LSH to introduce the local attention mechanism with the inductive bias of locality of convolution-like operators, expanding the reasonable receptive field, enhancing feature distinctiveness.
	
	\item To promote the information interaction of overlap regions within point cloud pairs, we propose the Interaction Transformer, which utilizes the overlap matrix computed from the global representation of each window to accurately match the overlap regions.
	
	\item Comprehensive experiments on real-world indoor and outdoor benchmarks demonstrate that LAHNet achieves significant registration results.

\end{itemize}

%\vspace{15pt}

\section{Related Works}

\subsection{Learning-based Point Cloud Descriptors}

%The formidable data-driven capabilities and trainable attributes of deep learning have significantly accelerated progress in the 3D field. Benefiting from the powerful expressive capacity of deep neural networks, directly learning features from 3D data has become feasible. This has prompted further development in learning-based point cloud descriptors for point cloud registration.

%The formidable data-driven capabilities and trainable attributes of deep learning have significantly accelerated progress in the 3D field. 

Leveraging the powerful representation capacity of deep neural networks, directly learning features from 3D data has become feasible \cite{qi2017pointnet, qu2024conditional, qi2017pointnet++, choy20194d}. This has driven the further development of learning-based point cloud descriptors for point cloud registration. 3DMatch \cite{zeng20173dmatch} takes the local geometry patches converted by point clouds as input, and learns geometry features by 3D Convolutional Neural Networks (3DCNNs). Meanwhile, similar to 3DMatch, SpinNet \cite{ao2021spinnet} transforms point clouds into a voxelized sphere and utilizes 3DCNNs to learn rotation invariant descriptors. However, although 3DCNNs can effectively extract local features from point clouds, this suffers from a significant computational cost. The pioneering work FCGF \cite{choy2019fully} utilizes 4D sparse convolutions \cite{choy20194d} to construct a fully convolutional network for learning local point cloud descriptors. This employs a U-Net network architecture to aggregate multi-scale local features, and utilizes residual connections to preserve more fine-grained information. The combination of sparse convolution and the U-Net network architecture significantly reduces the extraction time of point cloud descriptors. Follow this pattern, subsequent methods \cite{huang2021predator, horache20213d, huang2021imfnet, wang2023roreg} have achieved remarkable registration results. 

%PREDATOR \cite{huang2021predator} utilizes the attention mechanism to consider the overlap regions of point cloud pairs, and reaches a astonishing performance on low-overlap point cloud pairs. MS-SVConv \cite{horache20213d} employs a multi-scale architecture to learn local descriptors that exhibit robustness across different scales of point clouds.  

%Furthermore, IMFNet \cite{huang2021imfnet} designs a efficient Transformer module to fuse point clouds with images and achieves SOTA performance. 

Although these methods have achieved outstanding performance by focusing on optimizing local feature extraction, they fail to model long-range dependencies. This results in these methods still struggling with the feature ambiguity on ambiguous geometric structures, due to the limited feature receptive field. In this paper, we propose a novel point cloud descriptor that establishes reasonable long-range dependencies between points to expand the receptive field, enhancing feature distinctiveness.

\subsection{Point Cloud Transformers}
Transformer, with a powerful capability to model long-range dependencies, has achieved significant success in natural language processing and image vision. Inspired by this, PCT \cite{guo2021pct} pioneeringly  introduces Transformer into 3D tasks while retaining the global self-attention mechanism. However, the global attention may introduce redundant disturbing information \cite{xia2022vision} and cause feature averaging \cite{park2022how}, while the quadratic computational complexity poses scalability challenges. To address this issue, Point Transformer \cite{zhao2021point} adopts the U-Net network architecture, and performs self-attention within the local neighborhoods of each point. Meanwhile, Voxel Transformer \cite{mao2021voxel} applies Transformer to sparse voxels. Notably, Swin Transformer \cite{liu2021swin} introduces a windowed self-attention mechanism with inductive bias of locality. This partitions the image into non-overlapping regular windows and performs self-attention within each window with an offset, significantly reducing the computational cost. Motivated by this pattern, in contrast to Voxel Transformer, SST \cite{fan2022embracing} considers each voxel as a regular window to perform self-attention. However, the voxelization process itself can also introduce considerable computational overhead. Recently, some methods uniformly partition point clouds into non-overlapping windows by octrees, achieving remarkable results in 3D tasks \cite{wang2023octformer}. 

%Recently, some methods \cite{wu2022point, wu2024point} have focused on leveraging grid pooling to partition windows, achieving remarkable results.

% This focuses more on the local structure within each voxel rather than modeling the entire point cloud.

%However, the difference  of point number within each window requires the addition of pseudo-points to support parallelization, resulting in additional computational overhead. Flatformer \cite{liu2023flatformer} achieves the uniform distribution of voxelized points within each windows through a feature-flattening operation.

%Nevertheless, the voxelization itself also incurn significant computational costs.

In this paper, we leverage locality-sensitive hashing to introduce the local attention mechanism with inductive bias of locality into point cloud descriptors. This can uniformly partition the disordered  point cloud into regular and non-overlapping windows with linear computational costs.

%3DFeat-Net \cite{yew20183dfeat} learns local features with rotation invariance through a weakly supervised framework. Simultaneously,To improve the descriptor robustnes

\section{Locality-Sensitive Hashing}

In this section, we first briefly introduce locality-sensitive hashing (LSH), followed by our approach for window partitioning, along with an effectiveness analysis compared to voxelization, KNN and octree. Our goal is to present LSH as a supplement to the window partitioning strategy.

%along with the effectiveness analysis compared to KNN and voxelization.

\subsection{Background for LSH}
\label{3.1}

LSH leverages a predefined linear projection function $T(\cdot)$ to map points ($\bm{p_1}, \bm{p_2}$) from the original data space (Euclidean space) to a new data space (hashing domain):

\begin{equation}
    \label{f31}
    |T(\bm{p_1})-T(\bm{p_2})|\left\{
        \begin{array}{rcl}
        = 0, & \bm{p_1},\bm{p_2} \rightarrow adjcent\\
        > 0, & \bm{p_1},\bm{p_2} \rightarrow distant\\
        \end{array} \right.
\end{equation}

Eq.~\ref{f31} reveals two meaningful properties of LSH: (a) If $\bm{p_1}$ and $\bm{p_2}$ are adjacent in the original data space, then $T(\bm{p_1})=T(\bm{p_2})$; otherwise, $ T(\bm{p_1}) \neq T(\bm{p_2})$. (b) According to the linear function property, the more adjacent $\bm{p_1}$ and $\bm{p_2}$ are in the original data space, the closer $T(\bm{p_1})$ and $T(\bm{p_2})$ are. 

\begin{figure}[htp]
	\centering
	\includegraphics[width=0.48\textwidth]{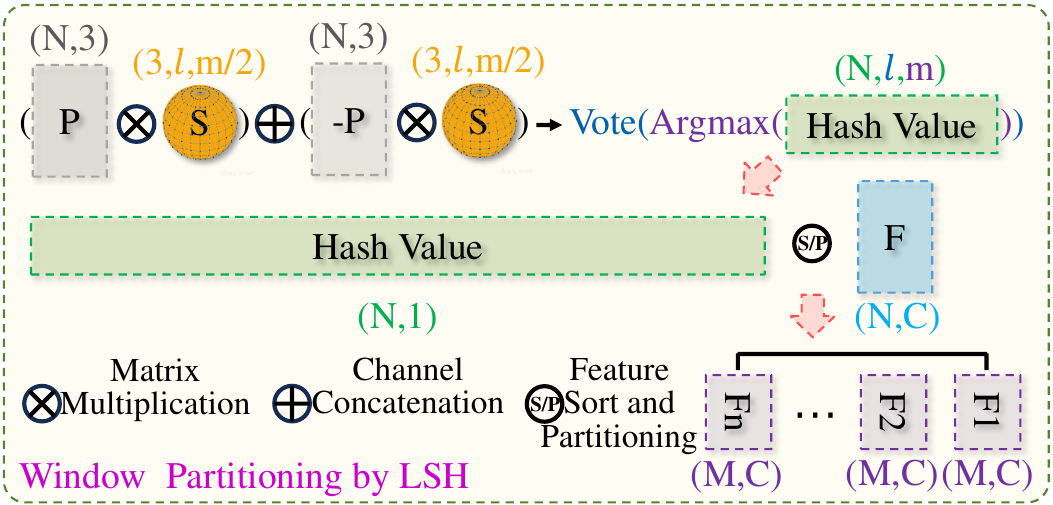}
	\caption{The window partitioning process by LSH. $S$ is a random rotation matrix following standard Gaussian distribution. $\bm{F}$ is uniformly partitioned into non-overlapping windows based on the hash values of $\bm{P}$ from $m$ bins.}
	\label{fig_LSH}
\end{figure}

\vspace{-15pt}

\subsection{Window Partitioning by LSH}
\label{3.2}

We apply the predefined hash function \cite{andoni2015practical} to achieve window partitioning. This can preserve the neighbor relationship between points in Euclidean space by projecting the original 3D points onto a unit sphere space (the linear function property, see Sec.~\ref{3.1}). We first obtains the corresponding hash values from point coordinates using LSH. Then, the features are sorted and partitioned by the corresponding hash values. The process is depicted in Fig.~\ref{fig_LSH}. 

Specifically, given the point set $\bm{P} \in \mathbb{R}^{N \times 3}$, the corresponding features $\bm{F} \in \mathbb{R}^{N \times C}$ and a randomly generated rotation matrix $S \in \mathbb{R}^{3 \times l \times m/2}$, we first obtain the corresponding hash values $\bm{H} \in \mathbb{R}^{N \times 1}$ of $\bm{P}$ from $m$ bins:

\vspace{-5pt}

\begin{equation}
    \label{f32}
    \bm{H}=vote(argmax(cat(\bm{P}S;-\bm{P}S)))
\end{equation}
where $cat(;)$ represents the channel concatenation. Meanwhile, $argmax(\cdot)$  
determines the bin index with the highest projected value (probability) as the corresponding hash value. Furthermore, $vote(\cdot)$ is a voting function that selects the most frequent hash value from $l$ rounds of LSH, mitigating the impact of randomness from $S$, enhancing the result robustness.

Then, $\bm{F}$ is sorted by $\bm{H}$ and sequentially uniformly partitioned into multiple non-overlapping windows:

\begin{equation}
    \label{f33}
    FI = partition(sort(\bm{F};\bm{H});n)
\end{equation}
where $FI=\{\bm{F_i}\in \mathbb{R}^{M \times C}|i=1..n\}$. $n$ and $M$ represents the number of windows and the number of points in each window, respectively. Meanwhile, $sort(;)$ means a sorting function. 

\begin{figure}[htp]
	\centering
	\includegraphics[width=0.495\textwidth]{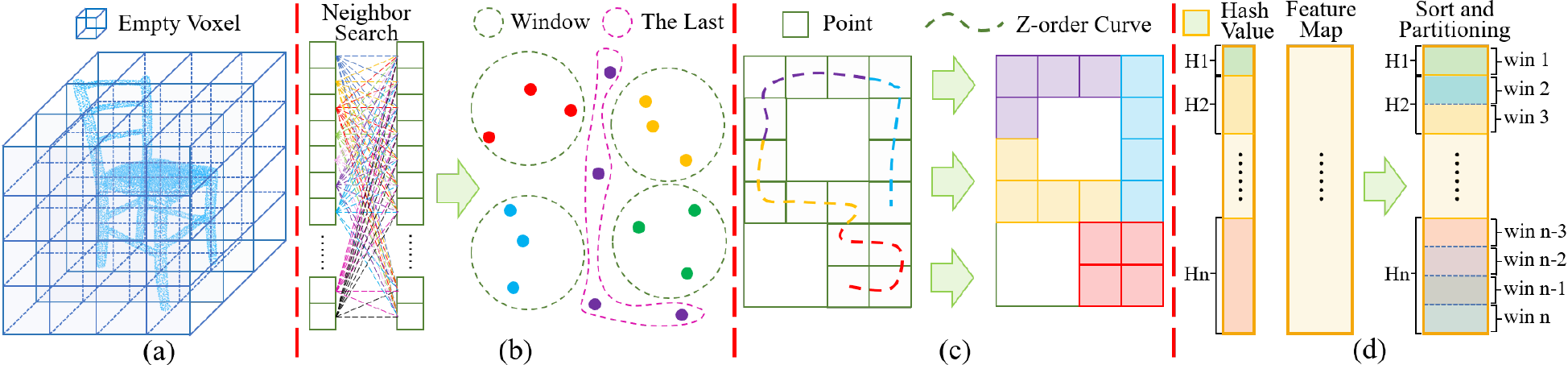}
	\caption{Voxelization vs. KNN vs. Octree vs. LSH. (a) Voxelization generates a large number of empty voxels, leading to additional computational costs. (b) KNN introduces quadratic complexity, while challenging to handle for the last window. (c) Octree constructs a z-order curve to uniformly partition the point cloud \cite{wang2023octformer}. (d) LSH can easily  partition windows based on hash values.}
	\label{fig_compare}
\end{figure}
\vspace{-10pt}

\subsection{Voxelization vs. KNN vs. Octree vs. LSH}

%To introduce the local attention mechanism, existing point cloud transformers often use voxelization and KNN for window partitioning \cite{zhao2021point, mao2021voxel, park2022fast, fan2022embracing, liu2023flatformer}. However, this either introduces redundant computational overhead or suffers from quadratic complexity. Fig \ref{fig_compare} illustrates this problem.

In Fig.~\ref{fig_compare}, we provide a preliminary comparison for partitioning point clouds into non-overlapping windows by voxelization, KNN, octree and LSH. (a) Although voxelization can regularly partition point clouds into regular cubic windows, considerable empty voxels lead to additional computational costs. Moreover, the voxelization process also incurs significant computational overhead. (b) KNN has quadratic complexity of neighborhood searches. Meanwhile, the last window is unreasonably formed by the remaining non-adjacent points. (c) The z-order curve is constructed through an n-depth octree  to uniformly partition windows based on segment lengths. (d) LSH can linearly project 3D points to hash values. According to Sec.~\ref{3.1}, we can uniformly partition point clouds into non-overlapping windows by sorting based on hash values, ensuring the reasonableness of the last window. Meanwhile, LSH is simpler to implement compared to an octree.

%  ($\bm{H_1}<\bm{H_2}<...<\bm{H_{n-1}}<\bm{H_n}$)

\begin{figure*}[htp]
	\centering
	\includegraphics[width=\textwidth]{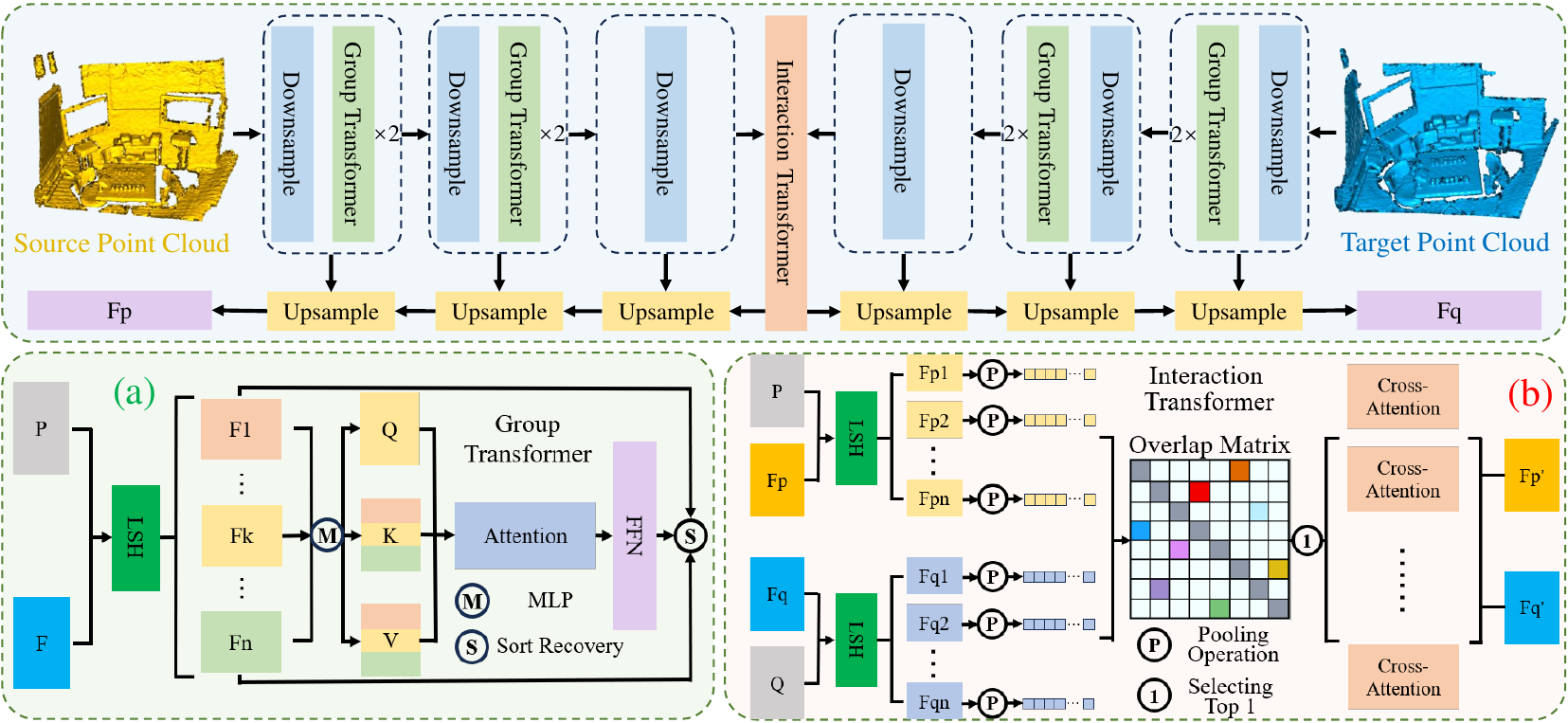}
	\caption{The overall architecture of LAHNet. The source point cloud and target point cloud are as inputs, and the output includes corresponding features, $\bm{F_p}$ and $\bm{F_q}$. First, two consecutive Group Transformers in the U-Net encoder are applied to model reasonable long-range dependencies between points. Next, a Interaction Transformer is placed at the U-Net bottleneck stage for feature interaction of overlap regions. Finally, the U-Net decoder focuses on feature scale restoration and aggregates multi-scale information.}
	\label{fig_network}
\end{figure*}

\section{Methodology}

%The overall network architecture is depicted in Fig \ref{fig_network}. Our method, adopting a local attention mechanism with a cross-window strategy to model the long-range dependencies between points, named LAHNet, applies a two-branch U-Net architecture. 

% In this section,, we introduce the network architecture  , as illustrated in Fig.~\ref{fig_network}. Our approach, which employs a local attention mechanism complemented by a cross-window strategy to capture the long-range dependencies between points, is named LAHNet. It utilizes a two-branch U-Net architecture.

In this section, we first provide the problem definition of point cloud registration, with the goal of finding a set of reliable correspondences through \textbf{distinctive features}. Then, we gradually introduce the overall framework of LAHNet, which  expands the reasonable receptive field by a local attention mechanism to learn \textbf{distinctive features}, as shown in Fig.~\ref{fig_network}.

\subsection{Problem Setting}
\label{4.1}
Given two partially overlapped point clouds $\bm{{P}} \in \mathbb{R}^{M \times 3}$ and $\bm{{Q}} \in \mathbb{R}^{N \times 3}$, the registration goal is to align $\bm{{P}}$ with $\bm{{Q}}$ by the estimated rigid transformation matrix $\bm{T}$ with the rotation matrix $\bm{R} \in SO(3)$ and the translation vector $\bm{t} \in \mathbb{R}^3$:

\begin{equation}
	\label{f41}
	\mathop{min}_{\bm{R} \in SO(3), \bm{t} \in \mathbb{R}^3} \sum_{(\bm{p_i},\bm{q_j}) \in C}||\bm{R}\bm{p_i} + \bm{t} - \bm{q_j}||_2^2
\end{equation}
where $\bm{R}$ and $\bm{t}$ can be estimated using an estimator (such as RANSAC and SVD) based on a set of corresponding points. In other words, it is crucial to find a set of reliable correspondences,  $C=\{(\bm{p_i},\bm{q_j})|i,j=1..n\}$, between point clouds pairs using  distinctive descriptors. 

\subsection{Downsample Layer and Upsample Layer}
\textbf{The downsample layer.} Similar to FCGF \cite{choy2019fully}, we use 4D sparse convolution \cite{choy20194d} to achieve downsampling. This includes batch normalization layers and ReLU activation functions, while preserving more fine-grained information through residual connections.

\textbf{The upsample layer.} The upsample layer adopts the same configuration as the downsample layer. Differently, the upsampling layers utilize 4D sparse deconvolution \cite{choy20194d} to restore feature scales, while aggregating multi-scale information from the downsample layers by long skip connections.

\subsection{Group Transformer}

To establish reasonable long-range dependencies between points, we propose the Group Transformer, adopting a local attention mechanism with a cross-window strategy (see Fig.~\ref{fig_network}(a)). 

\textbf{Local attention.} For the point set $\bm{P}$ and the corresponding features $\bm{F}$ within each window, we follow the standard self-attention design:

\vspace{-10pt}

\begin{equation}
\begin{split}
	\label{f42}
	\bm{O}=\bm{F}+MHSA(LN(\bm{F})) \\
        \bm{X}=\bm{O}+FFN(LN(\bm{O})) \;\;
\end{split}        
\end{equation}
where $LN(\cdot)$, $MHSA(\cdot)$ and $FFN(\cdot)$ represent the layer normalization, multi-head self-attention, and feed-forward network, respectively.

\begin{figure}[htp]
	\centering
	\includegraphics[width=0.45\textwidth]{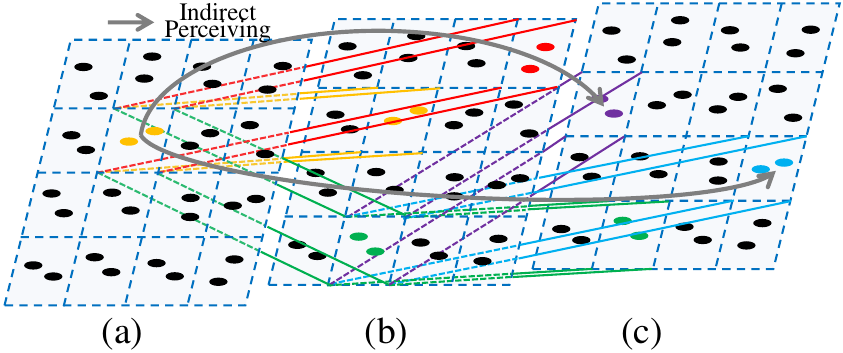}
	\caption{The process of enhancing feature receptive field through cross-window interaction. (a), (b), and (c) all represent the same point cloud partitioned into $4\times4$ windows.  The points (yellow) in (a) can indirectly perceive the information of points (purple and blue) in (c) through points (green) in (b).}
	\label{fig_cross-window}
\end{figure}

\textbf{Cross-window interaction.} The shifted window mechanism of Swin Transformer contributes to an enhanced receptive field for features. However, the mechanism is computationally expensive for point clouds, due to the unordered and irregular nature \cite{liu2023flatformer}. In this paper, we design a cross-window interaction strategy. Specifically, we keep $\bm{Q}$ unchanged within the current window, while combining points from the current window with points from randomly selected other windows to form $\bm{K}$ and $\bm{V}$ (see Fig.~\ref{fig_network}(a)). This simple and effective strategy can further expand the feature receptive field of the current window by overlaying that of other windows, as shown in Fig.~\ref{fig_cross-window}. We set the cross-window number to 2. 

% \textbf{The last window.} The difference of point scales in real datasets makes it challenging to ensure the consistent number of points in all last windows. For parallelization, we pad the number of points in the last window with $0$ to match the number of points in the previous windows. After the self-attention mechanism, we take out of the original points from the last window. This maintains the actual calculation results \cite{fan2022embracing}, and the additional computational cost from padding pseudo-points can be negligible. 

\vspace{-20pt}

\subsection{Interaction Transformer}

% to enhance the interaction of the overlap regions,

Benefiting from the simple and efficient window partitioning strategy (see Sec.~\ref{3.2}), we further propose an effective overlap region interaction module, Interaction Transformer (IT), which is formally based on multi-head cross-attention (see Fig.~\ref{fig_network}(b)). IT can learn an overlap matrix to match the overlap regions within point cloud pairs based on matching scores computed by global representations of windows. Unlike the global interaction that utilizes superpoint representations and requires additional overlap loss signals \cite{huang2021predator, qin2023geotransformer}, this focuses more on local overlap region matching.

\textbf{Overlap matrix.} Given the point sets $\bm{P} \in \mathbb{R}^{M \times 3}$ and $\bm{Q} \in \mathbb{R}^{N \times 3}$ from the encoder, and the corresponding features $\bm{F_{p}} \in \mathbb{R}^{M \times C}$ and $\bm{F_{q}} \in \mathbb{R}^{N \times C}$ respectively, we first utilize LSH to partition $\bm{F_{p}}$ and $\bm{F_{q}}$ into multiple uniform windows. Then, each window is transformed into a global representation by applying max-pooling over the spatial dimension to construct the overlap matrix (the elements in the overlap matrix represent the overlap ratio of the corresponding window pairs):

\vspace{-3pt}

\begin{equation}
\begin{split}
	\label{f43}
	\bm{G_p}=max(partition(sort(\bm{F_{p}};T(\bm{P}));n)) \\
        \bm{G_q}=max(partition(sort(\bm{F_{q}};T(\bm{Q}));n)) \\
        \bm{W}=\bm{G_p} \bm{G_q^T} \quad \quad \quad \quad \quad \quad \quad
\end{split}        
\end{equation}
where $\bm{G_p} \in \mathbb{R}^{n \times C}$, $\bm{G_q} \in \mathbb{R}^{n \times C}$ ($n$ indicates the window number). Meanwhile, $\bm{W} \in \mathbb{R}^{n \times n}$ represents the overlap matrix between windows of point cloud pairs. 

% where $\otimes$ means matrix multiplication. 

Subsequently, we match the overlap window pairs $\{(\bm{F_{p_i}},\bm{F_{q_j}})|i,j=1..n\}$ based on the overlap matrix $\bm{W}$. 

%However, the results may produce duplicated or missing windows due to the presence of non-overlapping regions between point cloud pairs. In the ablation experiments, we investigate the results of repeatable and unique selecting strategies.

\textbf{Overlap region interaction.} The points in the matching window pair ($\bm{F_{p_i}}  \in \mathbb{R}^{M' \times C}$, $\bm{F_{q_j}}  \in \mathbb{R}^{N' \times C}$) are interacted through the cross-attention mechanism:

\vspace{-3pt}

\begin{equation}
\begin{split}
	\label{f44}
	\bm{O}=\bm{F_{p_i}}+MHCA(LN(\bm{F_{p_i}}),LN(\bm{F_{q_j}})) \\
        \bm{X}=\bm{O}+FFN(LN(\bm{O})) \quad \quad \quad \;
\end{split}        
\end{equation}
where $MHCA(\cdot)$ means multi-head cross-attention.

Meanwhile, the interaction of matching window pairs is bidirectional, allowing the information to flow in both directions, $\bm{F_{p_i}} \rightarrow \bm{F_{q_j}}$, $\bm{F_{q_j}} \rightarrow \bm{F_{p_i}}$.

\subsection{Loss and Training}

\textbf{Contrastive loss.} We leverage the standard hardest-contrastive loss to learn the constraint of points, as the best performance is shown in FCGF \cite{choy2019fully}:

\begin{equation}
\begin{split}
	\label{f45}
	Loss = \sum_{i,j \in \mathcal{P}}\{[D(\bm{f_i},\bm{f_j})-0.1]_+^2 /|{\mathcal{P}}|\; \; \quad \quad \quad  \; \; \;\\
     + \; \; 0.5I_i([1.4-\mathop{min}_{k \in {\mathcal{N}_i}}D(\bm{f_i},\bm{f_k})]_+^2 /|{\mathcal{N}_i}|) \; \; \; \; \\
	+ \; \; 0.5I_j([1.4-\mathop{min}_{k \in {\mathcal{N}_j}}D(\bm{f_j},\bm{f_k})]_+^2 /|{\mathcal{N}_j}|)\} \;
\end{split}
\end{equation}
where ${\mathcal{P}}$ is a set of matched (positive) samples. Meanwhile, ${\mathcal{N}_i}$ and ${\mathcal{N}_j}$ are randomly sampled sets of non-matched (negative) points. $\bm{f_i}$, $\bm{f_j}$ and $\bm{f_k}$ mean the point features. $D(\cdot)$ represents the ${\mathcal{L}_2}$ distance. $I_i(\cdot)$/$I_j(\cdot)$ is an indicator function that returns 1 if $\bm{f_i}$/$\bm{f_j}$ and $\bm{f_k}$ form a negative pair, and 0 otherwise. $[\cdot]_+$ denotes the function $max(0,x)$.

This constrains positive pairs to be close to each other while keeping negative pairs far apart, enabling the model to learn distinctive features and achieve excellent correspondences between point cloud pairs.

\textbf{Training.} We train LAHNet for 200 epochs on an NVIDIA 3090 GPU, and the hyperparameter settings are followed FCGF \cite{choy2019fully}. For LSH, we set the number of bins $m=64$ and conduct $l=4$ rounds of voting. Meanwhile, for the local attention mechanism in GT and IT, we set the number of points in each window to $M=[128, 64]$ and $M=32$, the number of heads to $[2, 4]$ and $4$, and the dimension of each head to $[32, 64]$ and $128$, respectively.

\section{Experiments}

\subsection{Experiment Setup}
\textbf{Datasets.} Three public benchmarks, 3DMatch \cite{zeng20173dmatch}, 3DLoMatch \cite{huang2021predator} and KITTI \cite{geiger2012we}, are used for evaluation. For 3DMatch, we follow the official split strategy, 48 scenes for training, 6 scenes for validation and 8 scenes for testing. Meanwhile, for 3DLoMatch, we conduct training on 3DMatch and testing on the low-overlap counterpart of the 3DMatch test set \cite{huang2021predator}. Furthermore, for KITTI, we follow the approach in FCGF \cite{choy2019fully}, using 6 sequences for training, 2 sequences for validation, and 3 sequences for testing.

\textbf{Metrics.} We utilize several important metrics to evaluate the performance of point cloud descriptors \cite{yu2024riga}: feature matching recall (FMR), registration recall (RR), inlier ratio (IR), relative rotation error (RRE), and relative translation error (RTE).

\vspace{-3pt}

\begin{table}[h]
        \caption{Feature matching recall (FMR) on 3DMatch}
	\begin{center}
		%\scriptsize
		\begin{tabular}{p{2.5cm}|p{1.0cm}p{1.0cm}p{1.0cm}p{1.0cm}}	
        \Xhline{1pt}
		%\begin{tabular}{c|cccc}}	\hline
		Method
		&$\tau_2(5\%)$   
        &\makecell[c]{std$\downarrow$} 
		&$\tau_2(20\%)$   
        &\makecell[c]{std$\downarrow$}\\
		\hline
        
		FPFH\cite{rusu2009fast} 	             
		&\makecell[c]{36.4}           &\makecell[c]{13.6}
		&\makecell[c]{-}              &\makecell[c]{-}  \\

		3DMatch\cite{zeng20173dmatch} 	        
		&\makecell[c]{50.8}           &\makecell[c]{-}
		&\makecell[c]{4.3}            &\makecell[c]{-}\\
        
		FCGF\cite{choy2019fully}                
		&\makecell[c]{95.3}           &\makecell[c]{3.3}
		&\makecell[c]{67.4}           &\makecell[c]{-}\\
        
		D3Feat\cite{bai2020d3feat}              
		&\makecell[c]{95.5}           &\makecell[c]{3.5}
		&\makecell[c]{75.8}           &\makecell[c]{-}\\
        
		PREDATOR\cite{huang2021predator}        
		&\makecell[c]{96.7}           &\makecell[c]{-}
		&\makecell[c]{86.2}           &\makecell[c]{-}\\
        
		SpinNet\cite{ao2021spinnet}             
		&\makecell[c]{97.5}           &\makecell[c]{\underline{1.5}}
		&\makecell[c]{85.7}           &\makecell[c]{-}\\
        
		MS-SVConv \cite{horache20213d}           
		&\makecell[c]{\textbf{98.4}}  &\makecell[c]{-}
		&\makecell[c]{{89.9}}           &\makecell[c]{-}\\ 

		YOHO \cite{wang2022you}          
		&\makecell[c]{{98.2}}  
        &\makecell[c]{{1.5}}
		&\makecell[c]{\underline{90.9}}   
        &\makecell[c]{\underline{4.3}}\\ 

        RoReg \cite{wang2023roreg}        
		&\makecell[c]{{\underline{98.2}}}  &\makecell[c]{1.6}
		&\makecell[c]{{90.2}}   
        &\makecell[c]{{4.5}}\\ 

        RIGA \cite{yu2024riga}        
		&\makecell[c]{{\underline{98.2}}}  &\makecell[c]{-}
		&\makecell[c]{-}  
        &\makecell[c]{-}\\ 

        \rowcolor{gray!20}
		\textbf{LAHNet}
		&\makecell[c]{\textbf{98.4}}  &\makecell[c]{\textbf{1.4}}
		&\makecell[c]{\textbf{91.5}}  &\makecell[c]{\textbf{3.6}}\\
		\Xhline{1pt}
			
		\end{tabular}
	\end{center}
	
	\label{tab_3dmatch_fmr}
\end{table}

\vspace{-18pt}
 
\begin{table}[h]
        \caption{Feature matching recall (FMR) on 3DLoMatch}
	\begin{center}
		%\scriptsize
		\begin{tabular}{p{2.5cm}|p{1.0cm}p{1.0cm}p{1.0cm}p{1.0cm}}	
        
        \Xhline{1pt}
			%\begin{tabular}{c|cccc}}	\hline
		Method
		&$\tau_2(5\%)$   
        &\makecell[c]{std$\downarrow$} 
		&$\tau_2(20\%)$   
        &\makecell[c]{std$\downarrow$}\\
		\hline
        
		FCGF\cite{choy2019fully}                
		&\makecell[c]{76.6}           &\makecell[c]{\underline{0.06}}
		&\makecell[c]{41.0}           &\makecell[c]{0.09}\\
        
		D3Feat\cite{bai2020d3feat}              
		&\makecell[c]{67.3}           &\makecell[c]{\underline{0.06}}
		&\makecell[c]{27.8}           &\makecell[c]{\underline{0.07}}\\
        
		PREDATOR\cite{huang2021predator}        
		&\makecell[c]{78.6}           &\makecell[c]{0.08}
		&\makecell[c]{{43.9}}           &\makecell[c]{0.10}\\
        
		SpinNet\cite{ao2021spinnet}             
		&\makecell[c]{75.3}           &\makecell[c]{0.11}
		&\makecell[c]{42.2}           &\makecell[c]{0.13}\\
        
		MS-SVConv \cite{horache20213d}           
		&\makecell[c]{71.7}  		  &\makecell[c]{0.07}
		&\makecell[c]{32.4}           &\makecell[c]{0.08}\\ 
        
   	YOHO\cite{wang2022you}          
		&\makecell[c]{{{79.4}}}  		  &\makecell[c]{{0.09}}
		&\makecell[c]{{48.9}}           
        &\makecell[c]{{0.11}}\\

        RoReg \cite{wang2023roreg}        
		&\makecell[c]{{{82.1}}}  
        &\makecell[c]{{0.07}}
		&\makecell[c]{\underline{50.1}}   
        &\makecell[c]{{0.10}}\\ 

        RIGA \cite{yu2024riga}        
		&\makecell[c]{{\textbf{85.1}}}  &\makecell[c]{-}
		&\makecell[c]{-}  
        &\makecell[c]{-}\\ 

        \rowcolor{gray!20}
		\textbf{LAHNet}
		&\makecell[c]{\underline{83.3}}  
        &\makecell[c]{\textbf{0.05}}
		&\makecell[c]{\textbf{52.1}}  &\makecell[c]{\textbf{0.06}}\\
		\Xhline{1pt}
			
		\end{tabular}
	\end{center}

	\label{tab_3dlomatch_fmr}
\end{table}

\begin{figure*}[htp]
	\centering
	\includegraphics[width=\textwidth]{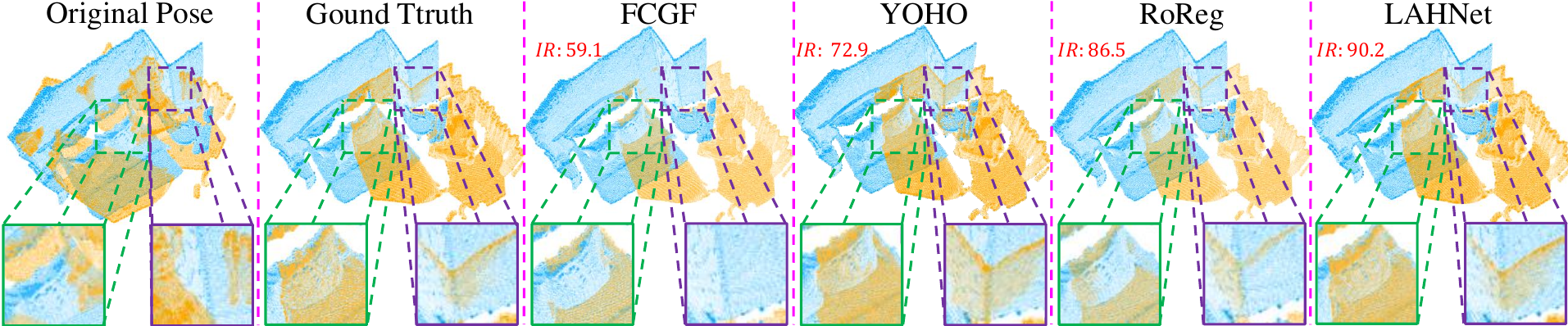}
	\caption{The visualization of registration results on 3DMatch. LAHNet achieves the high-quality registration result that exhibits the higher inlier ratio.}
	\label{fig_3dmatch}
\end{figure*}

\vspace{-15pt}

\subsection{Evaluation on 3DMatch and 3DLoMatch}
\textbf{Evaluation for the feature distinctiveness.} We first evaluate the feature distinctiveness of point cloud descriptors on indoor benchmarks. Tab.~\ref{tab_3dmatch_fmr} shows the results on 3DMatch. LAHNet consistently surpasses all methods. Notably, LAHNet achieves  better performance when $\tau_2$$\ge$$20\%$. This demonstrates that LAHNet is more robust than other descriptors under stricter inlier ratio thresholds. This is because, Group Transformer, by establishing reasonable long-range dependencies with the inductive bias of locality, enhances geometric awareness, thus improving feature distinctiveness. Meanwhile, Tab.~\ref{tab_3dlomatch_fmr} demonstrates that LAHNet also achieves significant performance on low-overlap 3DLoMatch when $\tau_2$$\ge$$20\%$, further proving the robustness under stricter thresholds.

\begin{table}[h]
        \caption{Different sampling robustness on 3DMatch}
	\begin{center}
		\scriptsize
		\begin{tabular}{p{1.8cm}p{0.7cm}p{0.7cm}p{0.7cm}p{0.7cm}p{0.7cm}p{0.6cm}}	
        \Xhline{1pt}
        
		\multirow{2}{*}{Method}   
		&\multicolumn{5}{c}{3DMatch \cite{zeng20173dmatch}}
		&\makecell[c]{\multirow{2}{*}{Avg$\uparrow$}} \\
		\cline{2-6} 
		&\makecell[c]{5000}
		&\makecell[c]{2500}
		&\makecell[c]{1000}
		&\makecell[c]{500}
		&\makecell[c]{250}
		& \\
		\hline
			
		\multicolumn{7}{c}{Feature Match Recall(FMR$\%$)$\uparrow$} \\
		\hline
			
		FCGF\cite{choy2019fully}         
		&\makecell[c]{95.2}    
        &\makecell[c]{95.5}   
        &\makecell[c]{94.6}
		&\makecell[c]{93.0}  
        &\makecell[c]{89.9}   
        &\makecell[c]{93.6}\\
        
%D3Feat(rand)\cite{bai2020d3feat}
%K&\makecell[c]{95.3}  
%&\makecell[c]{95.1}   
%&\makecell[c]{94.2}
%&\makecell[c]{93.6}    
%&\makecell[c]{90.8}   
%&\makecell[c]{93.8}\\	

		D3Feat\cite{bai2020d3feat}
		&\makecell[c]{95.8}    
        &\makecell[c]{95.6}     
        &\makecell[c]{94.6}
		&\makecell[c]{94.3}    
        &\makecell[c]{93.3}   
        &\makecell[c]{94.7}\\
        
		PREDATOR\cite{huang2021predator}
		&\makecell[c]{96.4}    	
        &\makecell[c]{96.3}    
        &\makecell[c]{96.0}
		&\makecell[c]{95.7}     
        &\makecell[c]{95.5}     
        &\makecell[c]{96.0}\\
        
		SpinNet\cite{ao2021spinnet}      
		&\makecell[c]{97.6}    
        &\makecell[c]{97.5}    
        &\makecell[c]{97.3}
		&\makecell[c]{96.3}     
        &\makecell[c]{94.3}     
        &\makecell[c]{96.6}\\ 
        
		MS-SVConv \cite{horache20213d}   
		&\makecell[c]{\textbf{98.4}}     
		&\makecell[c]{97.2}    
        &\makecell[c]{96.7}
		&\makecell[c]{96.4}     
        &\makecell[c]{93.7}    
        &\makecell[c]{96.5}\\ 

		YOHO \cite{wang2022you}   
		&\makecell[c]{{\underline{98.2}}}         
        &\makecell[c]{{{97.6}}}     
        &\makecell[c]{{{97.5}}}
		&\makecell[c]{{{97.7}}}         &\makecell[c]{{{96.0}}}    
        &\makecell[c]{{{97.4}}}\\ 

		RoReg \cite{wang2023roreg}   
		&\makecell[c]{{\underline{98.2}}}         &\makecell[c]{\underline{97.9}}     
        &\makecell[c]{\textbf{98.2}}
	&\makecell[c]{\underline{97.8}}         
        &\makecell[c]{{{97.2}}}     
        &\makecell[c]{\underline{97.9}}\\ 

		RIGA \cite{yu2024riga}   
		&\makecell[c]{{{97.9}}}        
        &\makecell[c]{{{97.8}}}     
        &\makecell[c]{{{97.7}}}
		&\makecell[c]{{{97.7}}}        
        &\makecell[c]{\textbf{97.6}}    
        &\makecell[c]{{{97.7}}}\\ 

        \rowcolor{gray!20}
		\textbf{LAHNet} 
		&\makecell[c]{\textbf{98.4}}        &\makecell[c]{\textbf{98.2}} 
		&\makecell[c]{\underline{97.8}}        	&\makecell[c]{\textbf{98.0}}
		&\makecell[c]{\underline{97.5}}        	&\makecell[c]{\textbf{98.0}}\\
		\hline
			
		\multicolumn{7}{c}{Registration Recall(RR$\%$)$\uparrow$} \\
		\hline
		FCGF\cite{choy2019fully}         
		&\makecell[c]{85.1}     
        &\makecell[c]{84.7}     
        &\makecell[c]{83.3}
		&\makecell[c]{81.6}     
        &\makecell[c]{71.4}    
        &\makecell[c]{81.2}\\
        
%D3Feat(rand)\cite{bai2020d3feat}
%&\makecell[c]{83.5}     
%&\makecell[c]{82.1}    
%&\makecell[c]{81.7}
%&\makecell[c]{77.6}    
%&\makecell[c]{68.8}     
%&\makecell[c]{78.7}\\		

		D3Feat\cite{bai2020d3feat}
		&\makecell[c]{82.2}     
        &\makecell[c]{84.4}     
        &\makecell[c]{84.9}
		&\makecell[c]{82.5}     
        &\makecell[c]{79.3}    
        &\makecell[c]{82.7}\\
        
		PREDATOR\cite{huang2021predator}
		&\makecell[c]{88.3}     
        &\makecell[c]{88.3}     
        &\makecell[c]{89.0}
		&\makecell[c]{88.4}     
        &\makecell[c]{\underline{84.7}}     &\makecell[c]{87.7}\\
        
		SpinNet\cite{ao2021spinnet}      
		&\makecell[c]{88.6}     
        &\makecell[c]{86.6}     
        &\makecell[c]{85.5}
		&\makecell[c]{83.5}     
        &\makecell[c]{70.2}     
        &\makecell[c]{82.9}\\ 
        
		MS-SVConv \cite{horache20213d}   
		&\makecell[c]{89.7}     
        &\makecell[c]{88.8}     
        &\makecell[c]{87.0}
		&\makecell[c]{82.6}     
        &\makecell[c]{73.8}     
        &\makecell[c]{84.4}\\ 

		YOHO \cite{wang2022you}   
		&\makecell[c]{{{90.8}}}        
        &\makecell[c]{{{90.3}}}     
        &\makecell[c]{{{89.1}}}
	&\makecell[c]{{{88.6}}}                       &\makecell[c]{{84.5}} 
        &\makecell[c]{{{88.7}}}\\ 

		RoReg \cite{wang2023roreg}   
		&\makecell[c]{\underline{92.9}}         &\makecell[c]{\underline{93.2}}     
        &\makecell[c]{\underline{92.7}}
	&\makecell[c]{\textbf{93.3}}         
        &\makecell[c]{\textbf{91.2}}     
        &\makecell[c]{\textbf{92.7}}\\ 

		RIGA \cite{yu2024riga}   
		&\makecell[c]{{{89.3}}}        
        &\makecell[c]{{{88.4}}}     
        &\makecell[c]{{{89.1}}}
		&\makecell[c]{{{89.0}}}        
        &\makecell[c]{{87.7}}    
        &\makecell[c]{{{88.7}}}\\ 

        \rowcolor{gray!20}
		\textbf{LAHNet} 
		&\makecell[c]{\textbf{94.2}}        &\makecell[c]{\textbf{93.4}} 
		&\makecell[c]{\textbf{93.2}}        &\makecell[c]{\underline{92.7}}
		&\makecell[c]{\underline{88.9}}        &\makecell[c]{\underline{92.3}}\\
		\hline
			
		\multicolumn{7}{c}{Inlier Ratio(IR$\%$)$\uparrow$} \\
		\hline
		FCGF\cite{choy2019fully}         
		&\makecell[c]{56.9}     
        &\makecell[c]{54.5}    
        &\makecell[c]{49.1}
		&\makecell[c]{43.3}    
        &\makecell[c]{34.7}    
        &\makecell[c]{47.7}\\
        
%D3Feat(rand)\cite{bai2020d3feat}
%&\makecell[c]{40.6}     
%&\makecell[c]{38.3}     
%&\makecell[c]{33.3}
%&\makecell[c]{28.6}     
%&\makecell[c]{23.5}    
%&\makecell[c]{32.9}\\

		D3Feat\cite{bai2020d3feat}
		&\makecell[c]{40.7}     
        &\makecell[c]{40.6}     
        &\makecell[c]{42.7}
		&\makecell[c]{44.1}    
        &\makecell[c]{45.0}    
        &\makecell[c]{42.6}\\
        
		PREDATOR\cite{huang2021predator}
		&\makecell[c]{49.9}     
        &\makecell[c]{50.3}     
        &\makecell[c]{49.2}
		&\makecell[c]{46.3}     
        &\makecell[c]{41.8}     
        &\makecell[c]{47.5}\\
        
		SpinNet\cite{ao2021spinnet}      
		&\makecell[c]{47.5}     
        &\makecell[c]{44.7}     
        &\makecell[c]{39.4}
		&\makecell[c]{33.9}     
        &\makecell[c]{27.6}     
        &\makecell[c]{38.6}\\ 
   
		MS-SVConv \cite{horache20213d}   
		&\makecell[c]{\underline{80.7}}          
        &\makecell[c]{\textbf{80.3}}     
        &\makecell[c]{\underline{76.1}}
		&\makecell[c]{{68.1}}          
        &\makecell[c]{{57.2}}     
        &\makecell[c]{{72.5}}\\ 

        YOHO \cite{wang2022you}   
        &\makecell[c]{{64.4}}              
        &\makecell[c]{{60.7}}     
        &\makecell[c]{{55.7}}
        &\makecell[c]{{46.4}}          
        &\makecell[c]{{41.2}}
        &\makecell[c]{{53.9}}\\ 

		RoReg \cite{wang2023roreg}   
		&\makecell[c]{{81.6}}         &\makecell[c]{{80.2}}     
        &\makecell[c]{{75.1}}
	&\makecell[c]{\underline{74.1}}         
        &\makecell[c]{\textbf{75.2}}     
        &\makecell[c]{\underline{77.3}}\\ 

		RIGA \cite{yu2024riga}   
		&\makecell[c]{{{68.4}}}        
        &\makecell[c]{{{69.7}}}     
        &\makecell[c]{{{70.6}}}
		&\makecell[c]{{{70.9}}}        
        &\makecell[c]{{71.0}}    
        &\makecell[c]{{{70.2}}}\\ 

        \rowcolor{gray!20}
		\textbf{LAHNet} 
		&\makecell[c]{\textbf{81.3}}        &\makecell[c]{\underline{80.0}} 
		&\makecell[c]{\textbf{79.5}}        &\makecell[c]{\textbf{78.4}}
		&\makecell[c]{\underline{72.9}}        &\makecell[c]{\textbf{78.4}}\\
		\Xhline{1pt}
        
		\end{tabular}
	\end{center}
	
	\label{tab_sample_3dmatch}
\end{table}

\begin{table}[h]
        \caption{Different sampling robustness on 3DLoMatch}
	\begin{center}
		\scriptsize
		\begin{tabular}{p{1.8cm}p{0.7cm}p{0.7cm}p{0.7cm}p{0.7cm}p{0.7cm}p{0.6cm}}	 
        \Xhline{1pt}
        
		\multirow{2}{*}{Method}   
		&\multicolumn{5}{c}{3DLoMatch \cite{huang2021predator}}
		&\makecell[c]{\multirow{2}{*}{Avg$\uparrow$}} \\
		\cline{2-6} 
		&\makecell[c]{5000}
		&\makecell[c]{2500}
		&\makecell[c]{1000}
		&\makecell[c]{500}
		&\makecell[c]{250}
		& \\
		\hline
			
		\multicolumn{7}{c}{Feature Match Recall(FMR$\%$)$\uparrow$} \\
		\hline
			
		FCGF\cite{choy2019fully}         
		&\makecell[c]{76.6}    
        &\makecell[c]{75.4}   
        &\makecell[c]{74.2}
		&\makecell[c]{71.7}    
        &\makecell[c]{67.3}  
        &\makecell[c]{73.0}\\
			
        D3Feat\cite{bai2020d3feat}
		&\makecell[c]{67.3}   
        &\makecell[c]{66.7}    
        &\makecell[c]{67.0}
		&\makecell[c]{66.7}   
        &\makecell[c]{66.5}   
        &\makecell[c]{66.8}\\	
        
		PREDATOR\cite{huang2021predator}
		&\makecell[c]{78.6}    
        &\makecell[c]{77.4}     
        &\makecell[c]{{76.3}}
		&\makecell[c]{{75.7}}      
        &\makecell[c]{{75.3}}  
        &\makecell[c]{{76.7}}\\
			
        SpinNet\cite{ao2021spinnet}      
		&\makecell[c]{75.3}    
        &\makecell[c]{74.9}    
        &\makecell[c]{72.5}
		&\makecell[c]{70.0}    
        &\makecell[c]{63.6}    
        &\makecell[c]{71.2}\\ 
        
		MS-SVConv \cite{horache20213d}   
		&\makecell[c]{71.7}    
        &\makecell[c]{70.5}    
        &\makecell[c]{69.4}
		&\makecell[c]{62.7}     
        &\makecell[c]{58.1}    
        &\makecell[c]{66.5}\\ 	

        YOHO \cite{wang2022you}   
		&\makecell[c]{{{79.4}}}    
        &\makecell[c]{{{78.1}}}     
        &\makecell[c]{{{76.3}}}
	&\makecell[c]{{73.8}}   
        &\makecell[c]{{69.1}}   
        &\makecell[c]{{75.3}}\\ 

		RoReg \cite{wang2023roreg}   
		&\makecell[c]{{{82.1}}}         &\makecell[c]{{82.1}}     
        &\makecell[c]{{81.7}}
	&\makecell[c]{{81.6}}         
        &\makecell[c]{{{80.2}}}     
        &\makecell[c]{{81.5}}\\ 

		RIGA \cite{yu2024riga}   
		&\makecell[c]{\textbf{85.1}}        
        &\makecell[c]{\textbf{85.0}}     
        &\makecell[c]{\textbf{85.1}}
		&\makecell[c]{\textbf{84.3}}        
        &\makecell[c]{\textbf{85.1}}    
        &\makecell[c]{\textbf{84.9}}\\ 

        \rowcolor{gray!20}
		\textbf{LAHNet} 
		&\makecell[c]{\underline{83.3}}        
        &\makecell[c]{\underline{83.2}} 
		&\makecell[c]{\underline{82.8}}       
        &\makecell[c]{\underline{82.5}}
		&\makecell[c]{\underline{82.0}}      
        &\makecell[c]{\underline{82.8}}\\
		\hline
			
		\multicolumn{7}{c}{Registration Recall(RR$\%$)$\uparrow$} \\
		\hline
		FCGF\cite{choy2019fully}         
		&\makecell[c]{40.1}  
        &\makecell[c]{41.7}    
        &\makecell[c]{38.2}
		&\makecell[c]{35.4}    
        &\makecell[c]{26.8}    
        &\makecell[c]{36.4}\\	
        
		D3Feat(\cite{bai2020d3feat}
		&\makecell[c]{37.2}     
        &\makecell[c]{42.7}    
        &\makecell[c]{46.9}
		&\makecell[c]{43.8}    
        &\makecell[c]{39.1}   
        &\makecell[c]{41.9}\\
        
		PREDATOR\cite{huang2021predator}
		&\makecell[c]{{59.8}} 
        &\makecell[c]{61.2}
        &\makecell[c]{62.4}
		&\makecell[c]{{60.8}}           
        &\makecell[c]{{58.1}}     
        &\makecell[c]{{60.5}}\\
        
		SpinNet\cite{ao2021spinnet}      
		&\makecell[c]{{59.8}}     
        &\makecell[c]{54.9}    
        &\makecell[c]{48.3}
		&\makecell[c]{39.8}    
        &\makecell[c]{26.8}     
        &\makecell[c]{45.9}\\ 
   
		MS-SVConv \cite{horache20213d}   
		&\makecell[c]{43.2}    
        &\makecell[c]{40.0}     
        &\makecell[c]{35.4}
		&\makecell[c]{31.3}     
        &\makecell[c]{22.2}     
        &\makecell[c]{34.4}\\ 
                
        YOHO \cite{wang2022you}   
		&\makecell[c]{{{65.2}}}    
        &\makecell[c]{{{65.5}}}     
        &\makecell[c]{{{63.2}}}
		&\makecell[c]{{56.5}}     
        &\makecell[c]{{48.0}}     
        &\makecell[c]{{59.7}}\\ 

		RoReg \cite{wang2023roreg}   
		&\makecell[c]{\textbf{70.3}}         &\makecell[c]{\textbf{71.2}}     
        &\makecell[c]{\textbf{69.5}}
	&\makecell[c]{\textbf{67.9}}         
        &\makecell[c]{\textbf{64.3}}     
        &\makecell[c]{\textbf{68.6}}\\ 

		RIGA \cite{yu2024riga}   
		&\makecell[c]{{65.1}}        
        &\makecell[c]{{64.7}}     
        &\makecell[c]{{64.5}}
		&\makecell[c]{{64.1}}        
        &\makecell[c]{{61.8}}    
        &\makecell[c]{{64.0}}\\ 

        \rowcolor{gray!20}
		\textbf{LAHNet} 
		&\makecell[c]{\underline{68.2}}       
        &\makecell[c]{\underline{67.9}} 
		&\makecell[c]{\underline{65.8}}      
        &\makecell[c]{\underline{64.5}}
		&\makecell[c]{\underline{63.0}}       
        &\makecell[c]{\underline{65.9}}\\
		\hline
			
		\multicolumn{7}{c}{Inlier Ratio(IR$\%$)$\uparrow$} \\
		\hline
		FCGF\cite{choy2019fully}         
		&\makecell[c]{21.4}    
        &\makecell[c]{20.0}     
        &\makecell[c]{17.2}
		&\makecell[c]{14.8}    
        &\makecell[c]{11.6}    
        &\makecell[c]{17.0}\\
        
		D3Feat\cite{bai2020d3feat}
		&\makecell[c]{13.2}    
        &\makecell[c]{13.1}   
        &\makecell[c]{14.0}
		&\makecell[c]{14.6}   
        &\makecell[c]{15.0}    
        &\makecell[c]{14.0}\\
        
		PREDATOR\cite{huang2021predator}
		&\makecell[c]{26.7} 
        &\makecell[c]{28.1}  
        &\makecell[c]{{28.3}}
		&\makecell[c]{{27.5}}    
        &\makecell[c]{{25.8}}   
        &\makecell[c]{{27.2}}\\
        
		SpinNet\cite{ao2021spinnet}      
		&\makecell[c]{20.5}    
        &\makecell[c]{19.0}    
        &\makecell[c]{16.3}
		&\makecell[c]{13.8}  
        &\makecell[c]{11.1} 
        &\makecell[c]{16.1}\\ 
        
		MS-SVConv \cite{horache20213d}   
		&\makecell[c]{{30.2}}     
        &\makecell[c]{{28.6}}     
        &\makecell[c]{25.7}
		&\makecell[c]{20.5}   
        &\makecell[c]{13.5}    
        &\makecell[c]{23.7}\\ 

        YOHO \cite{wang2022you}   
        &\makecell[c]{{25.9}}    
        &\makecell[c]{{23.3}}     
        &\makecell[c]{{22.6}}
        &\makecell[c]{{18.2}}     
        &\makecell[c]{{15.0}}     
        &\makecell[c]{{21.0}}\\ 

		RoReg \cite{wang2023roreg}   
		&\makecell[c]{\textbf{39.6}}         &\makecell[c]{\textbf{39.6}}     
        &\makecell[c]{{34.0}}
	&\makecell[c]{{31.9}}         
        &\makecell[c]{\underline{34.5}}     
        &\makecell[c]{\underline{35.9}}\\ 

		RIGA \cite{yu2024riga}   
		&\makecell[c]{{32.1}}        
        &\makecell[c]{{33.4}}     
        &\makecell[c]{\underline{34.3}}
		&\makecell[c]{\underline{34.5}}        
        &\makecell[c]{\textbf{34.6}}    
        &\makecell[c]{{33.8}}\\ 

        \rowcolor{gray!20}
        \textbf{LAHNet} 
        &\makecell[c]{\underline{39.2}}        	&\makecell[c]{\underline{38.1}} 
        &\makecell[c]{\textbf{37.5}}        	&\makecell[c]{\textbf{35.6}}
        &\makecell[c]{{32.5}}        	&\makecell[c]{\textbf{36.6}}\\
        
        \Xhline{1pt}
        
		\end{tabular}
	\end{center}
	
	\label{tab_sample_3dlomatch}
\end{table}

\textbf{Evaluation for different sampling robustness.} 
Subsequently, we evaluate the different sampling robustness of point cloud descriptors on 3DMatch and 3DLoMatch. Tab.~\ref{tab_sample_3dmatch} shows that LAHNet achieves superior results on IR. Benefiting from the long-range modeling capability with  inductive bias of locality, LAHNet can more accurately identify inliers, learning distinctive features. This is further demonstrated in Tab.~\ref{tab_sample_3dlomatch}, where LAHNet achieves remarkable results in inlier identification on the low-overlap 3DLoMatch.

Fig.~\ref{fig_3dmatch} visually shows the better registration performance of LAHNet compared to other methods on 3DMatch.

% In particular, LAHNet reaches $10.9\%$ and $4.5\%$ improvement for IR and RR, respectively, significantly outperforming other methods.

% Tab.\ref{tab_sample_3dlomatch} shows that LAHNet achieves $79.8\%$ ($\uparrow 3.1\%$) for FMR, and reaches $43.6\%$ ($\uparrow 16.4\%$) for IR.

\begin{figure*}[htp]
	\centering
	\includegraphics[width=\textwidth]{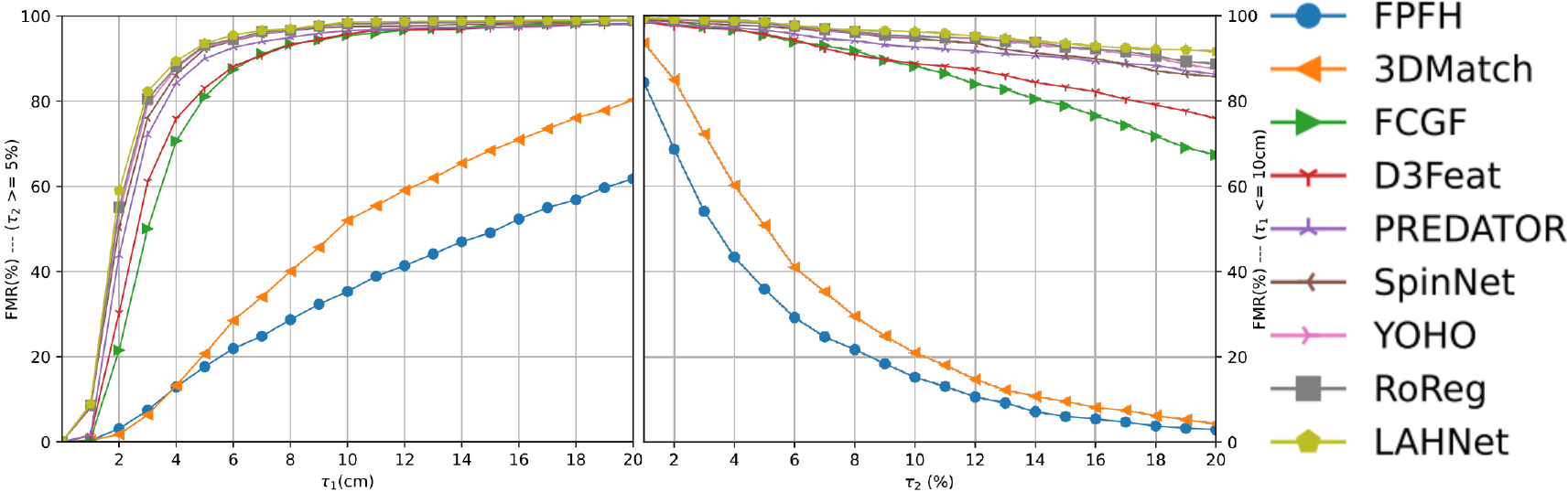}
	\caption{The results of different threshold settings (the default settings for FMR are $\tau_1$$\leq$$10cm$ and $\tau_2$$\ge$$5\%$ \cite{yu2024riga}) on 3DMatch. (a) The smaller $\tau_1$ ($\tau_2$$\ge$$5\%$), the more stringent the conditions, as $\tau_1$ means the inlier distance threshold between two points. (b) The larger $\tau_2$ ($\tau_1$$\leq$$10cm$), the more stringent the conditions, as $\tau_2$ indicates the inlier ratio threshold in two point clouds. LAHNet demonstrates more significant performance compared to other methods. }
	\label{fig_different_threshold}
\end{figure*}

\textbf{Evaluation for different threshold settings.}  Next, we investigate the performance of point cloud descriptors under different threshold settings. In Fig.~\ref{fig_different_threshold}, LAHNet consistently outperforms all methods, which further demonstrates the robustness and distinctiveness of LAHNet.

\begin{table}[h]
        \caption{Evaluation on KITTI}
        \vspace{-15pt}
	\begin{center}
		\scriptsize
		\begin{tabular}{p{2.3cm}|p{0.7cm}p{0.7cm}p{0.7cm}p{0.7cm}|p{1.2cm}}		
        \Xhline{1pt}
        
		\multirow{2}{*}{Method}  
		&\multicolumn{2}{c}{{RTE(cm)}} 
		&\multicolumn{2}{c|}{{RRE(°)}}
		&\makecell[c]{\multirow{2}{*}{{RR(\%)$\uparrow$}}}\\
		\cline{2-5}
        
		&\makecell[c]{Avg$\downarrow$} 
        &\makecell[c]{std$\downarrow$} 
        &\makecell[c]{Avg$\downarrow$} 
        &\makecell[c]{std$\downarrow$} \\  
		\hline
		
        FCGF\cite{choy2019fully}         
		&\makecell[c]{\underline{6.47}} 
        &\makecell[c]{1.30} 
        &\makecell[c]{\textbf{0.23}} 
        &\makecell[c]{0.23} 
        &\makecell[c]{98.9}\\
        
		D3Feat\cite{bai2020d3feat}       
		&\makecell[c]{6.90} 
        &\makecell[c]{\textbf{0.30}}
        &\makecell[c]{\underline{0.24}}
		&\makecell[c]{\textbf{0.06}}     &\makecell[c]{\textbf{99.8}}\\
        
		PREDATOR\cite{huang2021predator}     
		&\makecell[c]{6.80} 
        &\makecell[c]{-}    
        &\makecell[c]{0.27}
        &\makecell[c]{-}       
        &\makecell[c]{\textbf{99.8}}\\
        
		SpinNet\cite{ao2021spinnet}      
		&\makecell[c]{9.88} 
        &\makecell[c]{0.50}
        &\makecell[c]{0.47}
        &\makecell[c]{\underline{0.09}}     &\makecell[c]{99.1}\\

        RIGA \cite{yu2024riga}     
		&\makecell[c]{13.5} 
        &\makecell[c]{-}
        &\makecell[c]{0.45}
        &\makecell[c]{-}
        &\makecell[c]{99.1}\\
        
        \rowcolor{gray!20}
		\textbf{LAHNet} 
		&\makecell[c]{\textbf{6.40}} 
        &\makecell[c]{\underline{0.39}} 
		&\makecell[c]{0.26} 
		&\makecell[c]{0.19}
		&\makecell[c]{\underline{99.3}}\\
        
		\Xhline{1pt}
        
		\end{tabular}
	\end{center}
	
	\label{tab_Kitti}
\end{table}

\subsection{Evaluation on KITTI}

Furthermore, we also conduct the evaluation of point cloud descriptors on the outdoor benchmark. Tab.\ref{tab_Kitti} presents the results on KITTI. LAHNet demonstrates significant registration performance in the outdoor scene, particularly achieving remarkable results in terms of RTE. Compared to indoor scenes, the spatial distances between points are usually larger in outdoor point clouds, resulting in heightened data sparsity. This may require the model to build stronger point constraints. D3Feat \cite{bai2020d3feat} and PREDATOR \cite{huang2021predator} enhance the point constraints through additional keypoint detectors and overlap loss functions, respectively. In contrast, LAHNet learns an  overlap matrix to accurately match overlap regions of point cloud pairs through Interaction Transformer, perceiving reasonable correspondences between points in low-overlap and sparse scenarios.

\begin{figure*}[htp]
	\centering
	\includegraphics[width=\textwidth]{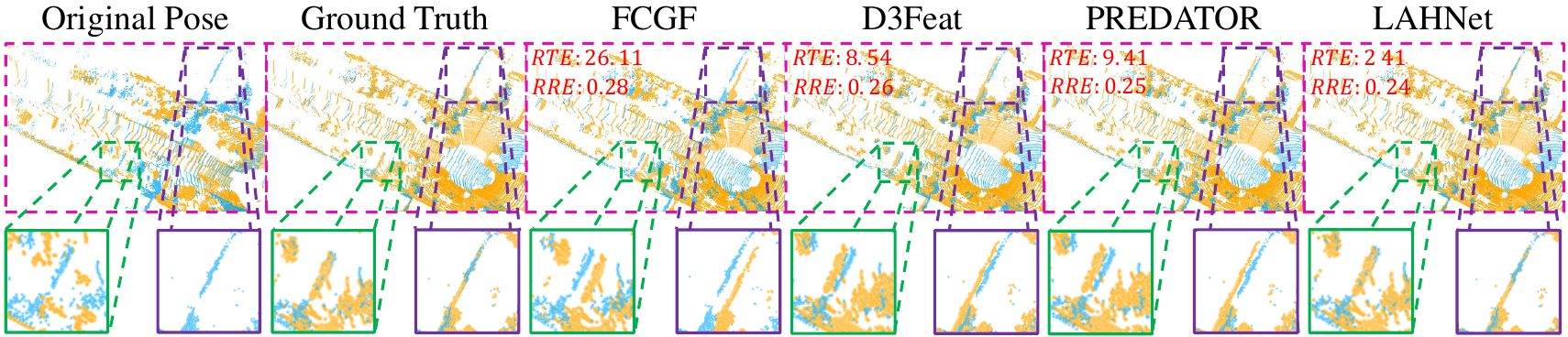}
	\caption{The visualization of registration results on KITTI. LAHNet demonstrates the outstanding registration result with the lower RTE and RRE
    .}
	\label{fig_kitti}
\end{figure*}

Fig.~\ref{fig_kitti} visualizes the registration results of LAHNet and other methods.

%Furthermore, we also conduct the evaluation of point cloud descriptors on the outdoor benchmark. Compared to indoor benchmarks, the point scale and density in KITTI is larger and sparser than 3DMatch, thus may requiring models with stronger point constraints. Tab.\ref{tab_Kitti} shows the results on KITTI. Admittedly, LAHNet is slightly lower than D3Feat \cite{bai2020d3feat} and PREDATOR \cite{huang2021predator}. The point scale and density in KITTI is larger and sparser than 3DMatch, thus may requiring models with stronger point constraints. D3Feat and PREDATOR enhance the point constraints respectively through additional keypoint detectors and loss functions. However, we strictly followed the training configuration of FCGF without adding additional point constraints, hence showing a slight disadvantage on KITTI. Moreover, in our experiments, we also observed that FCGF performs significantly better on 3DMatch, while performing poorly on KITTI.

\begin{table}[h]
	\caption{Ablation study for Group Transformer (GT) and Interaction Transformer (IT) on 3DMatch and 3DMatch}
    \vspace{-15pt}
	\begin{center}
        \scriptsize
		\begin{tabular}
        {p{0.65cm}p{0.65cm}|p{0.7cm}p{0.7cm}p{0.7cm}|p{0.7cm}p{0.7cm}p{0.7cm}}
		\Xhline{1pt}
  
        \makecell{\multirow{2}{*}{GT}}
        &\makecell{\multirow{2}{*}{IT}} 
        &\multicolumn{3}{c}{3DMatch($\%$)}
        &\multicolumn{3}{c}{3DLoMatch($\%$)} \\
        \cline{3-8}   
        
        &
        &\makecell[c]{FMR$\uparrow$}    
        &\makecell[c]{RR$\uparrow$}
        &\makecell[c]{IR$\uparrow$}   
        &\makecell[c]{FMR$\uparrow$}    
        &\makecell[c]{RR$\uparrow$}
        &\makecell[c]{IR$\uparrow$}                   \\
        
        \hline
        \makecell[c]{\XSolidBrush}
        &\makecell[c]{\XSolidBrush}
        &\makecell[c]{97.5}                 
        &\makecell[c]{89.9}
        &\makecell[c]{76.7}
        &\makecell[c]{77.5}                 
        &\makecell[c]{63.4}
        &\makecell[c]{34.2}              \\ 
        
        \makecell[c]{\Checkmark}
        &\makecell[c]{\XSolidBrush}           
        &\makecell[c]{\underline{98.1}}        
        &\makecell[c]{\underline{93.7}}
        &\makecell[c]{\underline{80.4}} 
        &\makecell[c]{\underline{81.5}}        
        &\makecell[c]{\underline{67.8}}
        &\makecell[c]{\underline{38.6}} \\ 
        
        \rowcolor{gray!20}
        \makecell[c]{\Checkmark}
        &\makecell[c]{\Checkmark}              
        &\makecell[c]{\textbf{98.4}}        
        &\makecell[c]{\textbf{94.2}}
        &\makecell[c]{\textbf{81.3}}
        &\makecell[c]{\textbf{83.3}}        
        &\makecell[c]{\textbf{68.2}}
        &\makecell[c]{\textbf{39.2}}   \\ 	
        
		\Xhline{1pt}	 
		\end{tabular}
	\end{center}

	\label{tab_GT_PIT}
\end{table}

\vspace{-10pt}

\subsection{Ablation Study}

\textbf{With/Without Group Transformer and  Interaction Transformer.} We verify the effectiveness of Group Transformer (GT) and Interaction Transformer (IT) in LAHNet on 3DMatch and 3DLoMatch. As shown in Tab.~\ref{tab_GT_PIT}, the registration performance is decreased significantly, when  Group Transformer and Interaction Transformer are removed from LAHNet. In this case (the first row in Tab.~\ref{tab_GT_PIT}), similar to most existing methods, this focuses on learning local features but lacks reasonable long-range dependency modeling. The limited receptive field causes a significant decrease in feature distinctiveness. Meanwhile, the lack of overlap region interaction makes features to hardly build high-quality correspondences between low-overlap point cloud pairs \cite{huang2021predator}.

% (in this case, LAHNet degenerates into FCGF)

%\textbf{KNN vs. LSH.} We conduct the ablation experiments for window partitioning using KNN and LSH. Tab \ref{tab_knn_lsh} demonstrates that LSH significantly surpasses KNN in descriptor extraction speed, while exhibiting better performance. 

%The quadratic complexity and the coarse processing for the last window make KNN significantly inferior to LSH in window partitioning, resulting in a significant decrease in descriptor extraction speed and performance.

\vspace{-5pt}
\begin{table}[h]
	\caption{Ablation study for rounds of LSH on 3DMatch}
	\begin{center}
        \scriptsize
		\begin{tabular}
        {p{1.0cm}p{1.2cm}p{1.0cm}p{1.0cm} p{2.1cm}}
		\Xhline{1pt}
  
		\makecell[c]{Rounds}  
		&\makecell[c]{FMR($\%$)$\uparrow$}    
        &\makecell[c]{RR($\%$)$\uparrow$}
        &\makecell[c]{IR($\%$)$\uparrow$}
		&\makecell[c]{FES(s/point)$\downarrow$}\\
        \hline  

        \makecell[c]{$l=1$}
        &\makecell[c]{98.3}                 
        &\makecell[c]{94.0}
        &\makecell[c]{81.1}
        &\makecell[c]{$9.25 \times 10^{-6}$}\\ 
        
        \makecell[c]{$l=2$}               
        &\makecell[c]{{98.3}}          
        &\makecell[c]{\underline{94.1}}
        &\makecell[c]{{81.0}} 
        &\makecell[c]{$9.29 \times 10^{-6}$}\\ 
        
        \rowcolor{gray!20}
        \makecell[c]{$l=3$}                 
        &\makecell[c]{\underline{98.4}}        
        &\makecell[c]{\textbf{94.2}}
        &\makecell[c]{\underline{81.2}}
        &\makecell[c]{$9.35 \times 10^{-6}$}\\ 	
        \rowcolor{gray!20}
        \makecell[c]{$l=4$}                 
        &\makecell[c]{\underline{98.4}}        
        &\makecell[c]{\textbf{94.2}}
        &\makecell[c]{\textbf{81.3}}
        &\makecell[c]{$9.41 \times 10^{-6}$}\\ 	

        \makecell[c]{$l=5$}                 
        &\makecell[c]{{98.3}}        
        &\makecell[c]{\underline{94.1}}
        &\makecell[c]{{81.1}}
        &\makecell[c]{$9.54 \times 10^{-6}$}\\ 	

        \makecell[c]{$l=6$}                 
        &\makecell[c]{\textbf{98.5}}        
        &\makecell[c]{\underline{94.1}}
        &\makecell[c]{\underline{81.2}}
        &\makecell[c]{$9.66 \times 10^{-6}$}\\ 	
        
		\Xhline{1pt}	 
		\end{tabular}
	\end{center}

	\label{tab_lsh}
\end{table}

\vspace{-20pt}

\begin{table}[h]
	\caption{Ablation study for cross-window number (CWN) on 3DMatch}
    \vspace{-15pt}
	\begin{center}
        \scriptsize
		\begin{tabular}
        {p{1.5cm}p{1.2cm}p{1.0cm}p{1.0cm} p{2.1cm}}
		\Xhline{1pt}
  
		\makecell[c]{Cross-Window}  
		&\makecell[c]{FMR($\%$)$\uparrow$}    
        &\makecell[c]{RR($\%$)$\uparrow$}
        &\makecell[c]{IR($\%$)$\uparrow$}
		&\makecell[c]{FES(s/point)$\downarrow$}\\
        \hline  

        \makecell[c]{$CWN=0$}
        &\makecell[c]{98.1}                 
        &\makecell[c]{93.9}
        &\makecell[c]{81.0}
        &\makecell[c]{$9.01 \times 10^{-6}$}\\ 
        
        \makecell[c]{$CWN=1$}               
        &\makecell[c]{{98.2}}          
        &\makecell[c]{{94.0}}
        &\makecell[c]{{81.1}} 
        &\makecell[c]{$9.23 \times 10^{-6}$}\\ 
        
        \rowcolor{gray!20}
        \makecell[c]{$CWN=2$}                 
        &\makecell[c]{\underline{98.4}}        
        &\makecell[c]{\textbf{94.2}}
        &\makecell[c]{\textbf{81.3}}
        &\makecell[c]{$9.41 \times 10^{-6}$}\\ 	
        \rowcolor{gray!20}
        \makecell[c]{$CWN=3$}                 
        &\makecell[c]{\textbf{98.5}}        
        &\makecell[c]{\underline{94.1}}
        &\makecell[c]{\textbf{81.3}}
        &\makecell[c]{$9.59 \times 10^{-6}$}\\ 	

        \makecell[c]{$CWN=4$}                 
        &\makecell[c]{\underline{98.4}}        
        &\makecell[c]{\textbf{94.2}}
        &\makecell[c]{\underline{81.2}}
        &\makecell[c]{$9.75 \times 10^{-6}$}\\  	
        
		\Xhline{1pt}	 
		\end{tabular}
	\end{center}

	\label{tab_cwn}
\end{table}

\vspace{-5pt}

\textbf{The rounds of LSH.}  We conduct ablation experiments for rounds of LSH, as the randomness introduced by $S$ may lead to non-adjacent points with the same hash value. Tab.~\ref{tab_lsh} shows that 3 or 4 rounds of LSH achieves the best trade-off between performance and feature extraction speed (FES). Meanwhile, as the rounds of LSH increases, the performance has reached saturation. This means that 3 or 4 rounds of LSH are sufficient to mitigate the randomness introduced by $S$. Moreover, the stable FES demonstrates that  the efficient matrix multiplication operation ensures that multiple rounds of LSH incur minimal computational cost.

\textbf{The cross-window number.} We further conduct ablation experiments for the cross-window number (CWN) in Group Transformer. As shown in Tab.~\ref{tab_cwn}, additional window interactions can further improve the performance. Meanwhile, 2 or 3 additional windows achieve the best trade-off between performance and efficiency. Moreover, we can observe in the experiments that as the number of merged windows further increases, the performance tends to saturate. This indicates that 2 or 3 window interactions   provide a sufficient feature receptive field. 

% which is similar to the shifted window mechanism in Swin Transformer, further enhancing the feature receptive field.

\vspace{-5pt}

\section{Conclusion}

In this paper, we first analyzed the effectiveness of a reasonable and broader (non-global) receptive field in learning distinctive descriptors for point cloud registration. Subsequently, we proposed Group Transformer, which models reasonable long-range dependencies between points through the local attention mechanism using LSH. Meanwhile, similar to the shifted window mechanism of Swin Transformer, we employed a cross-window interaction strategy to further enhance the feature receptive field. Furthermore, leveraging this efficient window partitioning strategy, we proposed  Interaction Transformer to enhance the feature interaction of the overlap regions within point cloud pairs. On real-world indoor and outdoor benchmarks, we demonstrated that our method achieves remarkable registration performance.

\vspace{-5pt}

{
    \small
    \bibliographystyle{IEEEtran}
    \bibliography{IEEEfull}
}

\vfill

\end{document}